\documentclass[letterpaper, 10 pt, conference]{ieeeconf}  

\IEEEoverridecommandlockouts                              

\usepackage{graphics} 
\usepackage{epsfig} 
\usepackage{mathptmx} 
\usepackage{times} 
\usepackage{amsmath} 
\usepackage{amssymb}  
\usepackage{booktabs}
\usepackage{multirow}
\usepackage{bbding}
\usepackage{makecell}
\usepackage{color}
\usepackage{amsfonts}
\usepackage{mathtools}
\usepackage{subcaption}

\DeclareMathAlphabet{\mathcal}{OMS}{cmsy}{m}{n}
\usepackage{algorithm}
\usepackage{algpseudocode}
\usepackage{hyperref}

\newcommand{\etal}{\textit{et al.}}

\usepackage{authblk}

\title{\LARGE \bf
ODIP: Towards Automatic Adaptation for Object Detection by\\ Interactive Perception
}

\author{Tung-I Chen$^1$, Jen-Wei Wang$^1$, and Winston H. Hsu$^{1, 2}$ 

\thanks{\\$^1$ National Taiwan University, 1, Section 4, Roosevelt Road, Taipei 10617\\
        $^2$ Mobile Drive Technology\\
        }%

}

\begin{document}

\maketitle
\thispagestyle{empty}
\pagestyle{empty}

\begin{abstract}
Object detection plays a deep role in visual systems by identifying instances for downstream algorithms.
In industrial scenarios, however, a slight change in manufacturing systems would lead to costly data re-collection and human annotation processes to re-train models.
Existing solutions such as semi-supervised and few-shot methods either rely on numerous human annotations or suffer low performance.
In this work, we explore a novel object detector based on interactive perception (ODIP), which can be adapted to novel domains in an automated manner.
By interacting with a grasping system, ODIP accumulates visual observations of novel objects, learning to identify previously unseen instances without human-annotated data.
Extensive experiments show ODIP outperforms both the generic object detector and state-of-the-art few-shot object detector fine-tuned in traditional manners.
A demo video is provided to further illustrate the idea~\cite{demo2021video}.

\end{abstract}

\section{INTRODUCTION}
%
%

%
For automated manufacturing systems and warehouses, the cycle time of product transition could be short.
To accommodate different workflows, it is important to develop a flexible robotic system that can generalize to novel domains within a short period.
In a common robotic system, the upstream algorithms, such as object detectors~\cite{ren2015faster, lin2017focal}, aim to provide visual perception for downstream robotic agents.
The more adaptable the upstream algorithm is, the more flexible downstream robotic systems could be.
In this research, we explore a novel object detector interacting with a grasping system, which can generalize to previously unseen objects without any human-annotated data.
\begin{figure}[t!]
    \centering
    \includegraphics[width=\linewidth]{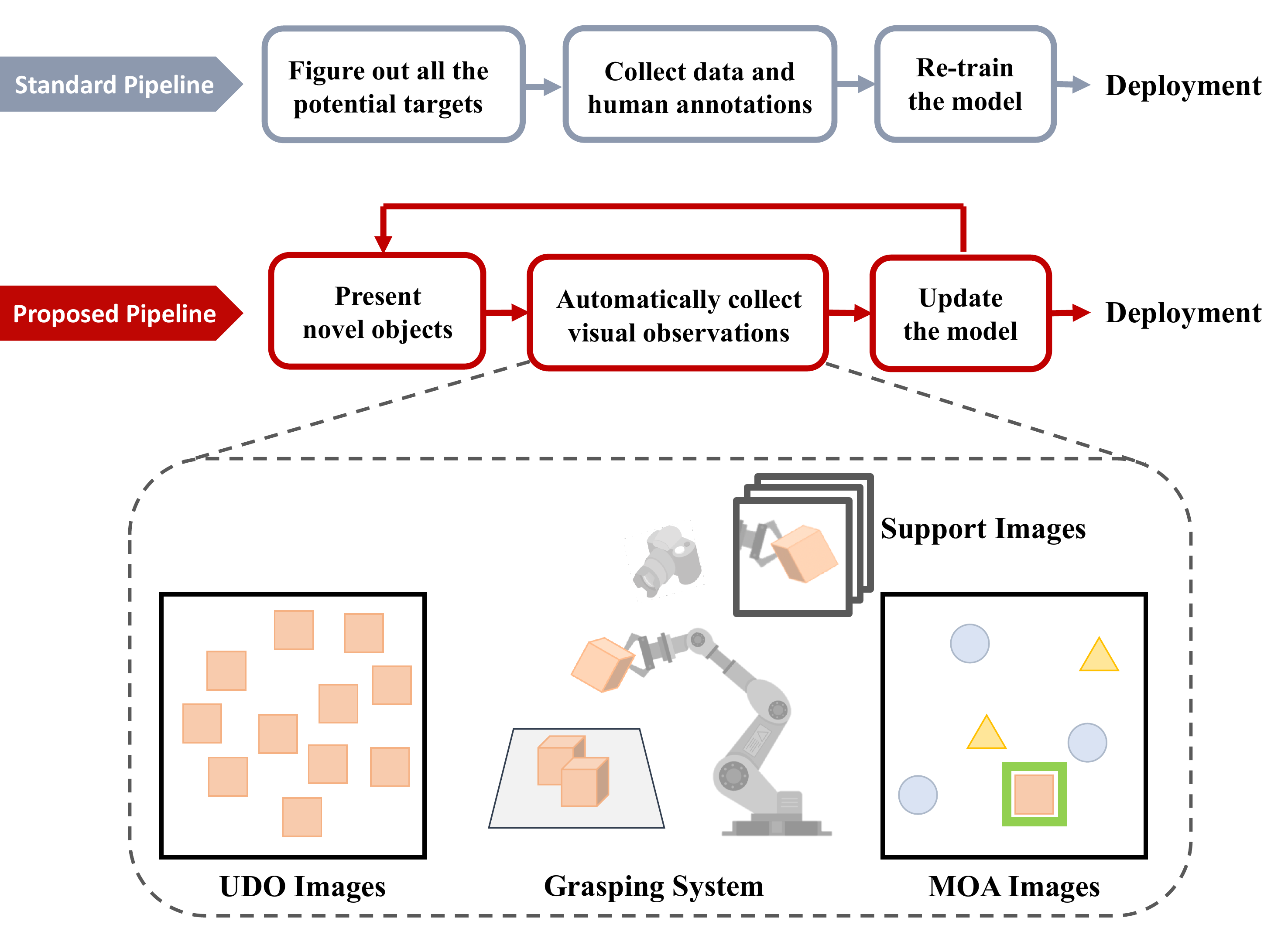}
    \caption{The comparison between the standard fine-tuning procedure and the proposed method.
    Generally, a considerable amount of annotated data must be re-collected to accommodate new workflows, which could be time-consuming and expensive in real-world scenarios.
    In this work, we propose a novel object detector, ODIP, which generalizes to previously unseen domains by merely interacting with a grasping system.
    Whenever novel objects are presented, ODIP can collect and learn visual observations in an automated manner without human-annotated data.
    }
    \label{fig:fig1}
\end{figure}
%

%
%

%
To develop a flexible deep learning-based object detector, the lack of annotated data is always the major challenge.
As illustrated in Fig.~\ref{fig:fig1}, to follow traditional fine-tuning procedures, numerous annotated data would be required, which could be costly in real-world situations.
In addition, since the deployed model can only recognize the objects included in the fine-tuning set, users must figure out all the potential targets in the upcoming workflows before re-collecting data.
While few-shot object detection (FSOD)~\cite{fan2020few, fan2020fgn} has made recent progress in alleviating data dependency, the performance of FSOD models is significantly inferior to conventional object detection models~\cite{ren2015faster, lin2017focal}.
Without a large corpus of annotated data, none of the existing methods reaches promising performance.
Observe these challenges, we propose a novel object detector called \emph{object detection by interactive perception} (ODIP), which learns to identify novel objects by leveraging the interactions between a FSOD model~\cite{chen2021should} and an object-agnostic grasping system~\cite{jeng2020coarse}.
Instead of treating fine-tuning as a static process, the domain transition of ODIP is performed in an automated and online manner.
There is no need to presume the targets of the upcoming workflows, and models can be facilitated without human intervention.
%
%

%
The basic idea of ODIP is to automatically collect visual observations via an object-agnostic grasp detector~\cite{qin2020s4g, jeng2020coarse}, which can infer viable grasps based on point clouds irrespective of object categories.
To train ODIP, the prerequisites are as follows: multiple novel (unseen) objects will be placed on the novel table (N-table), and a few base (seen) objects will be placed on the base table (B-table).
The robotic arm is asked to (1) pick up a novel object from the N-table, and (2) show the grasped object to a camera in different views, and then (3) place it on the B-table (as illustrated in the demo video~\cite{demo2021video}).
In each iteration, the scenes of the N-table can be captured as \emph{unlabeled, dense-object} (UDO) images, and the scenes of the B-table are captured as \emph{multi-class, one-shot annotated} (MOA) images.
During the interaction, the objects on the B-table must be sparse (less than six objects in our experiments) to prevent physical collisions between objects.
Without collision, the 3D coordinates of the novel object placed on the B-table can be inferred by the robotic configuration during the release.
The system can therefore acquire one-shot bounding box annotations of MOA images (see Fig.~\ref{fig:fig3}.~(a)).
Training ODIP by MOA images with only sparse target objects would inhibit models from tackling cluttered environments.
The multi-class, dense-object and fully-annotated data, however, are not available without human labeling.
Motivated by self-training techniques~\cite{xie2020self, zoph2020rethinking}, we propose a joint-training strategy to address the difficulty.
With class-agnostic FSOD models, the system can leverage unlabeled UDO images by assigning predicted labels to them.
The quality of such pseudo labels might be a concern in the early stages, but it can be significantly improved as the system collects more data to update the model.
By leveraging both MOA and UDO data, ODIP is capable of identifying instances within complicated environments (see Fig.~\ref{fig:prediction}).
\begin{figure}[t!]
    \centering
    \includegraphics[width=\linewidth]{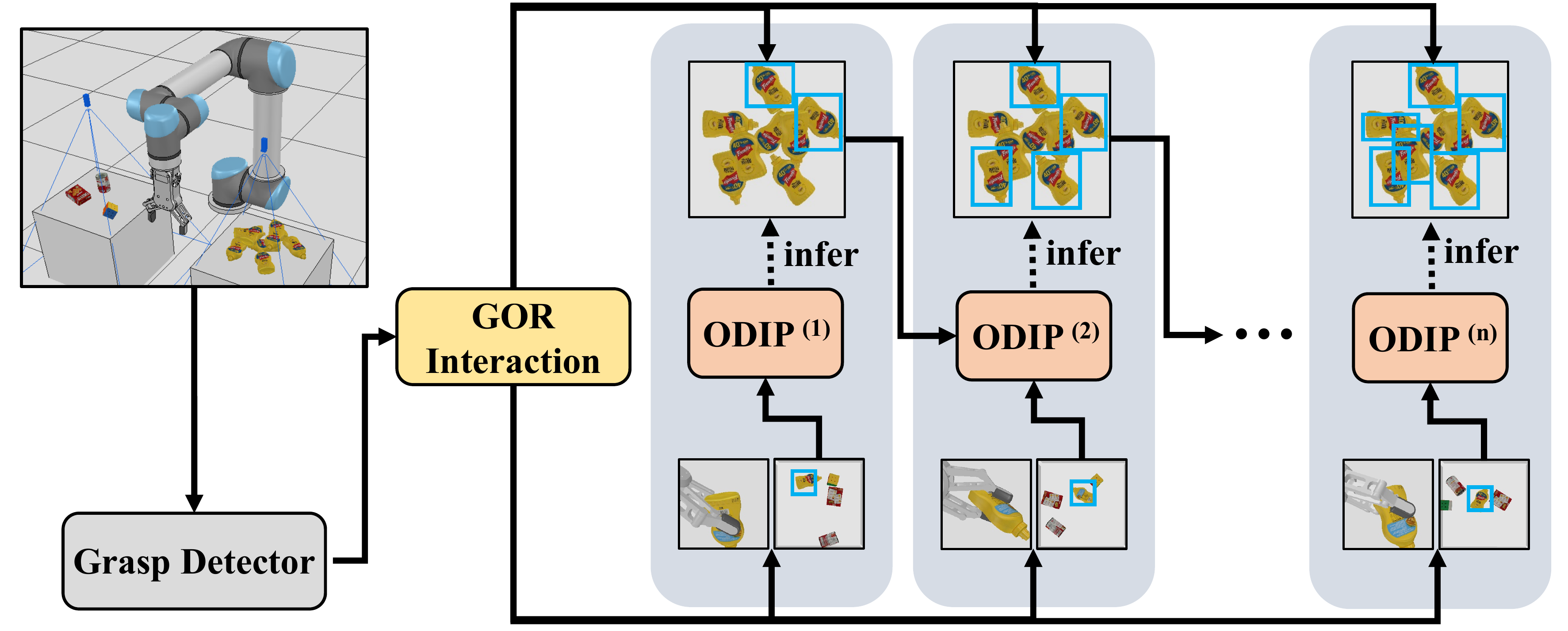}
    \caption{Overview of the proposed ODIP. The interaction between FSOD models and grasping systems can engender the required images (unlabeled, partially labeled, and sample images) for adaptation. The pseudo annotations of unlabeled images can be gradually refined as more visual observations are collected.
    }
    \label{fig:fig2}
\end{figure}
%
%
%

%
%
%
%

%
All the experiments in this work are carried out in a robotic simulator due to the efficiency and reproducibility of robotic experiments.
To test generalization ability, we leverage the simulator and YCB object set~\cite{calli2015ycb} to generate \emph{YCB2D} and \emph{Dense-YCB2D} datasets.
The proposed YCB2D includes 4 object categories (Fig.~\ref{fig:prediction}).
%
%
To train ODIP, we simulate a real-world scenario of workflow transition where the target objects are changing over time.
Objects of different categories are sequentially presented to the system, and the system has to gradually recognize those objects.
The baseline methods, on the other hand, can only be fine-tuned with fully annotated data prepared in advance.
Though ODIP is trained under a more challenging setting, it still achieves promising results in comparison to the traditional object detector (Tab.~\ref{tab:stage}).
%
%

%

%
Overall, our contributions can be summarized as follows:
\begin{itemize}
    \item We propose a novel object detector interacting with a grasping system, which can be updated in an automated and online manner.
    \item We propose a joint-training strategy, efficiently exploiting both unlabeled and partially-labeled data collected by the interaction.
    \item In comparison to existing methods, our approach achieves superior performance without fully-labeled data and is capable of tackling cluttered objects.  
  
\end{itemize}
\begin{figure}[t!]
    \centering
    \includegraphics[width=\linewidth]{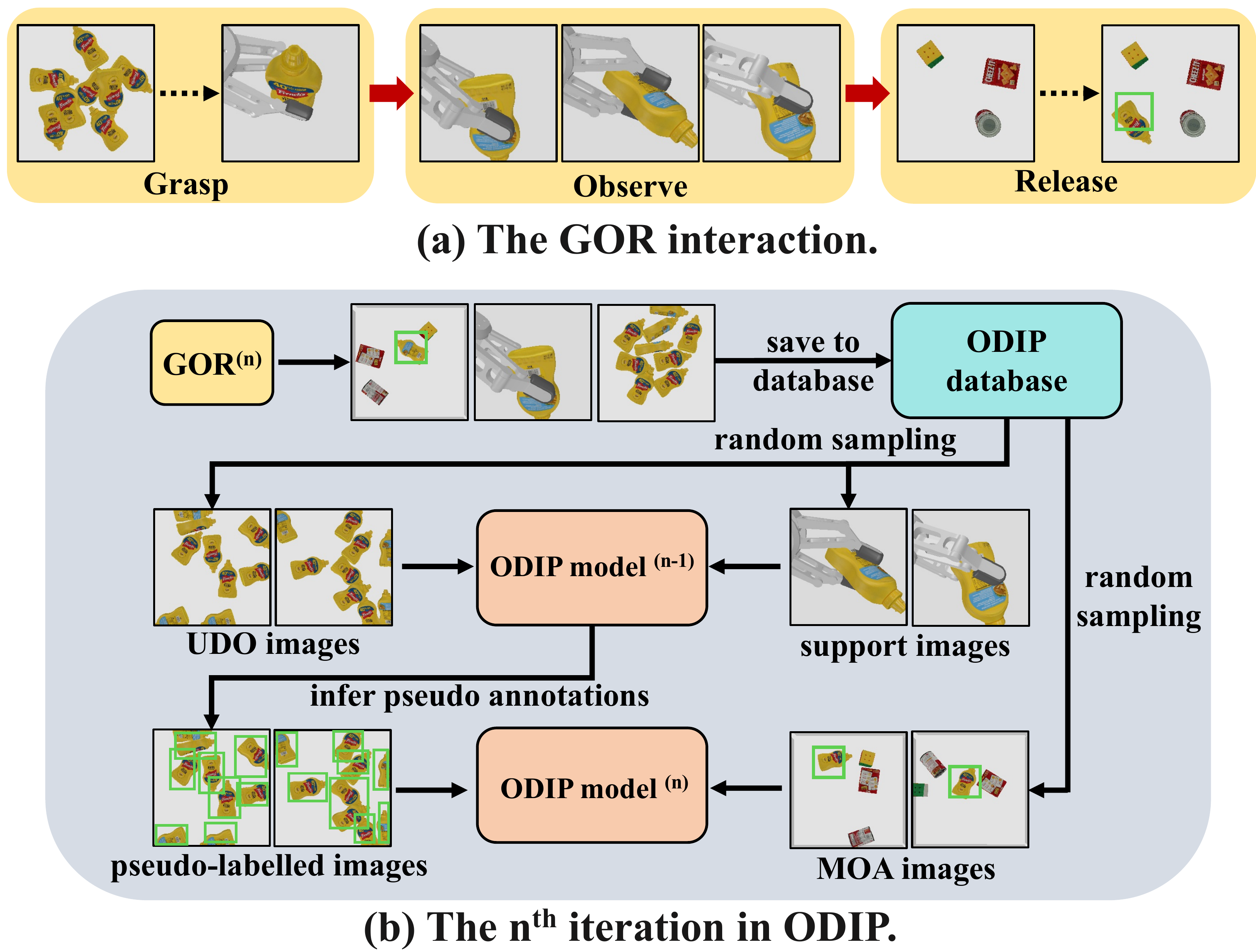}
    \caption{(a) The proposed grasp-observe-release interaction aims to engender the scenes which can be captured as source data for domain transition. (b) The system leverages class-agnostic FSOD models to assign pseudo labels to unlabeled images, and the predicted labels would be improved as more visual observations are collected. By training on both multi-class and dense-object images, ODIP shows remarkable ability even confronts cluttered environments.
    }
    \label{fig:fig3}
\end{figure}
\section{RELATED WORK}
\subsection{Semi-supervised Learning and Self-training}
%
%
%
The costly human annotation process usually limits us from leveraging most of the real-world data.
Semi-supervised learning (SSL) and self-training techniques have become increasingly popular in the research community.
Transductive label propagation~\cite{iscen2019label} is one of the important branches of SSL, which infers pseudo labels based on manifold assumption~\cite{iscen2019label}.
The idea of knowledge distillation~\cite{hinton2015distilling} is widely adopted in SSL as well, where a teacher model transfers its knowledge to students by providing them with predicted pseudo labels~\cite{yalniz2019billion}. 
Xie \etal~\cite{xie2020self} demonstrated that fitting both labeled and pseudo-labeled data can result in better performance, and the idea is further applied to object detection by Zoph \etal~\cite{zoph2020rethinking}, showing that leveraging ImageNet~\cite{deng2009imagenet} as a bounteous self-training data source can reach remarkable results on challenging COCO benchmark.
%
%

%
\subsection{Few-shot Object Detection}
%
Traditional object detectors~\cite{ren2015faster, lin2017focal} can only identify instances included in the training data, seriously restricting their potential applications.
Recent studies have explored few-shot object detection (FSOD) that facilitates models to generalize to novel domains with limited labeled images.
\cite{karlinsky2019repmet} proposed RepMet based on distance metric learning (DML), detecting novel objects by measuring the feature distances with prototypes in embedding spaces.
Apart from DML, another important line of work is based on attention mechanisms~\cite{fan2020fgn, fan2020few}, which employed attention to guide the detection networks to locate and classify instances.
%
Most FSOD approaches would adopt the strategies that have been proved effective in FSC, such as global pooling and feature map concatenation~\cite{sung2018learning}.
However, object detection is a much more challenging task than classification.
Ignoring the issues of spatial misalignment and biased samples could impair the performance of FSOD models.
Observe the challenges, \cite{chen2021should} proposed dual-awareness attention to precisely guide detection networks, and the proposed model manifested remarkable generalization ability compared to prior methods~\cite{fan2020fgn, fan2020few}.
Though the community has made significant progress in FSOD, the low performance would inhibit existing methods from being applied to the applications requiring high accuracy and stability. 
To achieve competitive performance, the size of a training dataset is still one of the most crucial factors.
%

%
\subsection{6-DoF Grasp Detection}
%
The 6-DoF grasp is promising in many real-world applications because of the dexterity to deal with cluttered environments.
Recent works are capable of directly inferring grasps from point cloud without assuming a known CAD model~\cite{ten2017grasp, liang2019pointnetgpd}.
The two-stage grasp detectors would be less efficient due to the onerous sampling and evaluating processes.
To improve efficiency, the end-to-end frameworks have been proposed~\cite{ni2020pointnet++, qin2020s4g} which can predict grasp configurations and corresponding quality scores in a single stage.
The methods based on cascaded framework~\cite{murali20206} are capable of collision-free grasping in dense clutter with the aid of instance segmentation model cropping the point cloud of specific objects.
However, it remains challenging to completely encode the diversity of grasp distributions on clutter scenes.
Therefore, Jeng \etal~\cite{jeng2020coarse} proposed a coarse-to-fine (C2F) method to enrich the diversity, improving the accuracy of multi-trial grasps significantly.
In our experiments, we leveraged the C2F grasping model~\cite{jeng2020coarse} to build the grasping system due to its high stability and accuracy.

%

%
\section{Object Detection by Interactive Perception}
\label{section:method}
\begin{algorithm}[t!]
\caption{Object Detection by Interactive Perception}
\label{alg:algorithm1}
\begin{algorithmic}[1]
\Require a grasp detector $g_{\theta}$, a few-shot object detector $f_{\theta_{0}}$, $T$, $N$, $L$, $\eta$
\State Initialize $\mathcal{D}_{UDO}\gets\{\}$, $\mathcal{D}_{MOA}\gets\{\}$, $\mathcal{D}_{Support}\gets\{\}$
\For{each stage $t=1,2,...,T$}
    \For{$n=1,2,...,N$}
        \State $\zeta$ $\gets$ Reset the environment
        \State $\tilde{x}_{i},s_{i},x_{i},z_{i}\gets$ $GOR(g_{\theta}, \zeta)$
        \State $\mathcal{D}_{UDO}\gets \mathcal{D}_{UDO} + \{\tilde{x}_{i}\}$
        \State $\mathcal{D}_{Support}\gets \mathcal{D}_{Support} + \{s_{i}\}$
        \State $\mathcal{D}_{MOA}\gets \mathcal{D}_{MOA} + \{(x_{i}, z_{i})\} $
        
    \EndFor
    \State Sample $\mathcal{S}\gets\{s_{i}\}$ from $D_{support}$
    \State Infer pseudo labels $\{\tilde{z}_{i}\}_{i=1}^{t\times N} \gets f_{\theta_{t-1}}(\mathcal{D}_{UDO}| \mathcal{S})$
    \State Define $\mathcal{D}_{Pseudo}\gets\{(\tilde{x}_{i}, \tilde{z}_{i})\}_{i=1}^{t\times N}$
    \State Define $\mathcal{D}_{All}\gets\mathcal{D}_{Pseudo}\oplus D_{MOA}$
    \State Let $f_{\theta_{t}}\gets f_{\theta_{0}}$
    \State Sample task set $\mathcal{T}$ from $\mathcal{D}_{All}$ and $D_{Support}$ to fine-tune $f_{\theta_{t}}$ until convergence
    \For{$l=1,2,...,L$}
        \State Sample task set $\mathcal{T}$ from $\mathcal{D}_{MOA}$ and $D_{Support}$
        \State Fine-tune $f_{\theta_{t}}$ with learning rate $\eta$
    \EndFor
    \State Evaluate $f_{\theta_{t}}$
\EndFor

\end{algorithmic}
\end{algorithm}
%

\subsection{Preliminary}
The basic idea of ODIP is to collect visual data and labels by the interaction between a grasping system and a few-shot object detector.
To train ODIP, a well-developed FSOD model is indispensable since we will have to infer pseudo annotations of previously unseen instances as novel objects are presented.
In the following sections, we will briefly introduce the problem setting of FSOD and then explain ODIP in detail.
Let $x$ denote a query image (image to be tested) and $s$ denote a support image (sample).
To perform FSOD, the model $f_{\theta}(x|\mathcal{S})$ should identify the instances in $x$ contingent upon a given support set
$\mathcal{S}=\{s\}_{i=1}^n$.
Unlike conventional object detection, a foreground instance is regarded as a target only if its category is consistent with the given support set.
We will use $z$ to denote those bounding box annotations of positive instances.
For FSOD, the basic unit of a dataset is a meta-learning task $\mathcal{T}=\{(x, S, z)\}$ comprised of queries, supports and corresponding box annotations. 
To train or evaluate a model, a task set $\{\mathcal{T}\}_{i=1}^{n}$ should be constructed in advance.
%
%

\subsection{Grasp-Observe-Release Interaction}
The proposed method is illustrated in Algorithm~\ref{alg:algorithm1}.
As novel objects are presented, the grasping system will be asked to perform a series of actions, which is called \emph{Grasp, Observe} and \emph{Release} (GOR) interaction in this work.
A batch of query and support images can be collected by performing GOR, and the annotations of images can be either estimated by robotic configurations or inferred by the model trained in the last stage.
To reset the environment, the system can directly remove all the objects on the B-table and re-scatter some base objects.
Since we have to record the coordinates of the released object, the base objects placed on the B-table must be sparse to prevent physical collisions.
For the N-table, there is no such restriction but all the novel objects on it should belong to the same category.
The point cloud information $\zeta$ of objects on the N-table will be captured by a commodity RGBD camera and sent to the grasp detection model $g_{\theta}$.
Note that the employed grasp detector should be capable of inferring viable grasps based on partial point cloud from a single view, such as \cite{jeng2020coarse}. 
As illustrated in Fig.~\ref{fig:fig3}.~(a), the GOR interaction consists of three actions. (1) The robotic arm picks up an object from the N-table according to the predicted grasps. (2) The mounted camera captures the images of the grasped object in different views. (3) The grasped object is released on the B-table.
If the grasp in (1) fails, we can just restart the process.
The scenes of the N-table are captured as unlabeled, dense-object (UDO) query images $\tilde{x}_i$, and the scenes of the B-table are captured as multi-class, one-shot annotated (MOA) query image $x_i$.
The close-up photos obtained in (2) will be the support images $s_i$ for FSOD models.
Generally, the ground-truth box annotations of query images are unavailable without human labeling.
However, the robot configurations enable the system to compute the 3D coordinates of that object released in (3), which will be the only novel object on the B-table due to the environment settings.
Thus, a coarse, one-shot bounding box label of the novel object in $x_i$ can be estimated.
Though the annotations of UDO images $\tilde{x}_i$ could not be directly inferred by interaction, the system can still leverage them by self-training approaches.
\begin{table}[t!]
    \centering
    \resizebox{\linewidth}{!}{
        \begin{tabular}{c | c|c c c c}
            \toprule
                \multirow{2}{*}{Stage} & \multirow{2}{*}{Added Category} & Cube & Can & Box & Bottle\\
                \cline{3-6}
                && \multicolumn{4}{c}{$AP$} \\
                \hline
                \multirow{4}{*}{Stage $1$} & Cube & 46.5 & - & - & - \\
                 & Can & 47.4 & 23.2 & - & - \\
                 & Box & 48.0 & 23.0 & 45.3 & - \\
                 & Bottle & 47.9 & 23.7 & 45.9 & 22.4 \\
                \bottomrule
        \end{tabular}
    }
    \caption{In the experiments, different novel objects are sequentially introduced to the system and the model must identify those objects that have been presented. ODIP addresses the issue of catastrophic forgetting by managing a database to preserve the collected data. Note that this table includes the first-stage results only, and the performance can be improved as more visual data are collected. 
}
    \label{tab:further_illustration}
\end{table}

\subsection{Joint-training Strategy}
\label{tab:disentangling}
After the novel instances in MOA images have been annotated, the system has already obtain necessary data to update FSOD models since the other base objects in the images will be considered as backgrounds contingent upon the given support set. 
However, training on MOA with only sparse objects would restrict models from tackling complicated scenes, which could be verified by our ablation study (Tab.~\ref{tab:ablation}).
Motivated by self-training~\cite{xie2020self, zoph2020rethinking}, we propose a joint-training strategy that enables ODIP to tackle cluttered environments without human annotation.  
%
%

\begin{figure}[t!]
    \centering
    \includegraphics[width=\linewidth]{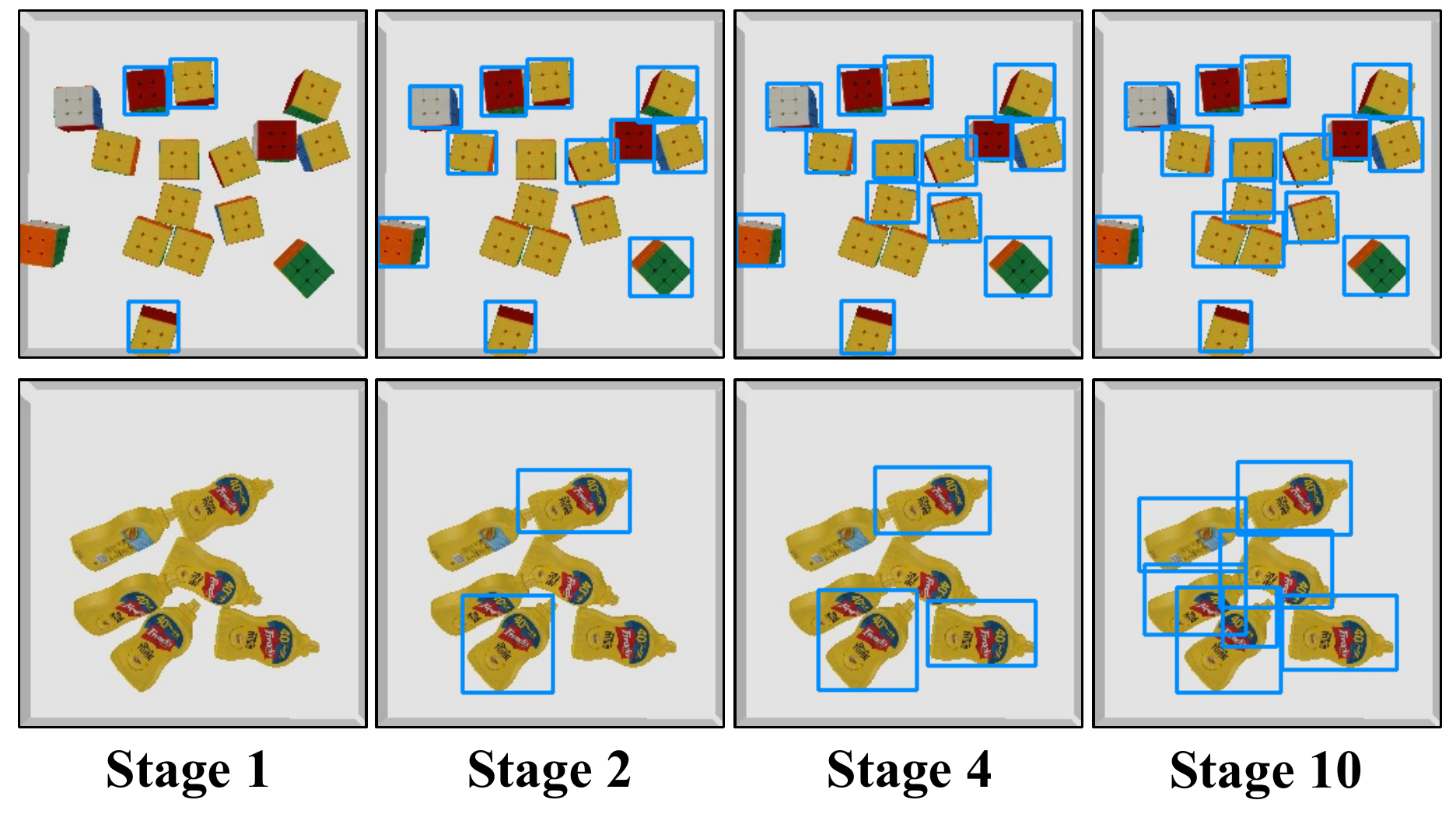}
    \caption{Visualization of the predicted labels in different stages. The predicted bounding box annotations of UDO images can be refined as the stage increases.
    }
    \label{fig:pseudo}
\end{figure}

\begin{figure*}[t!]
    \centering
    \includegraphics[width=\linewidth]{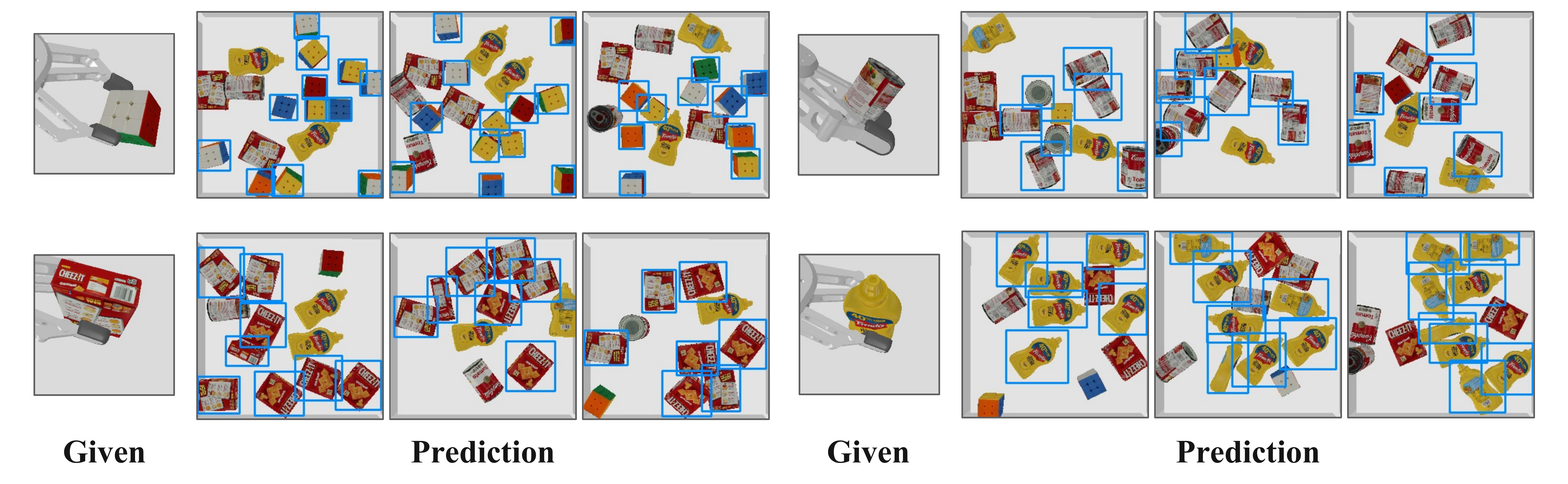}
    \caption{Visualization of the predictions of YCB2D. ODIP is capable of recognizing specific objects in cluttered environments contingent upon given sample images. Note that ODIP does \textbf{NOT} leverage any human annotation during the adaptation.   
    }
    \label{fig:prediction}
\end{figure*}

%
Ideally, models should be trained on the images comprised of dense, multi-class objects with precise bounding box annotations, but such data are unavailable without human intervention. 
Therefore, one of the expedient measures would be disentangling the ideal data into two different domains, the dense-object UDO images and the multi-class MOA images.
Both of them can be collected during the GOR interaction.
The labels of MOA can be estimated by the grasping system, while the labels of UDO can be inferred by the FSOD model $f_{\theta_{t-1}}$.
Note that the quality of pseudo labels in UDO could be improved by collecting more data to update $f_{\theta}$ (see Fig.~\ref{fig:pseudo}).
To fine-tune ODIP, a task set $\{\mathcal{T}\}$ will be sampled from the joint-training set $\mathcal{D}_{All}$, which is comprised of both UDO and MOA images.
Since the pseudo annotations might contain considerable noise in early stages, ODIP will be slightly fine-tuned on only MOA data before proceeding to the next stage or being deployed~\cite{yalniz2019billion}.
%

%
%
%
%

%
\begin{table*}[t!]
    \centering
    \resizebox{\linewidth}{!}{
        \begin{tabular}{c|c c| c c c c|c c|c c c c}
            \toprule
                \multirow{3}{*}{Method} & \multicolumn{6}{c|}{YCB2D dataset} & \multicolumn{6}{c}{Dense-YCB2D dataset} \\
                \cline{2-13}
                & \multicolumn{2}{c|}{All} & Cube & Can & Box & Bottle & \multicolumn{2}{c|}{All} & Cube & Can & Box & Bottle \\
                \cline{2-13}
                & \multicolumn{1}{c}{$AP$} & \multicolumn{1}{c|}{$AP_{50}$} & \multicolumn{4}{c|}{$AP$} & \multicolumn{1}{c}{$AP$} & \multicolumn{1}{c|}{$AP_{50}$} & \multicolumn{4}{c}{$AP$} \\
                \hline
                Faster R-CNN $256$~\cite{ren2015faster} & 46.9 & 80.9 & 54.5 & 43.3 & 46.8 & 43.1 
                & 31.8 & 62.9 & 46.8 & 41.8 & 24.7 & 21.1 \\
                Faster R-CNN $512$~\cite{ren2015faster} & 52.0 & 81.9 & 60.7 & 46.2 & 52.9 & 48.1 
                & 34.7 & 64.4 & 48.3 & 42.4 & 25.1 & 22.8 \\
                Faster R-CNN $1024$~\cite{ren2015faster} & 53.7 & 83.1 & 65.8 & 47.4 & 56.8 & 44.8 
                & 35.9 & 64.6 & 55.4 & 43.4 & 26.0 & 18.7 \\
                Faster R-CNN $2048$~\cite{ren2015faster} & 54.0 & 83.6 & 63.6 & 46.1 & 57.0 & 49.5 
                & 35.8 & 66.4 & 50.9 & 42.6 & 27.2 & 20.4 \\
                \hline
                DAnA $40$~\cite{chen2021should} & 29.2 & 55.1 & 45.0 & 23.1 & 28.2 & 20.7 
                & 18.6 & 41.9 & 30.4 & 19.0 & 15.0 & 10.1 \\
                DAnA $200$~\cite{chen2021should} & 47.0 & 78.5 & 57.1 & 39.5 & 54.5 & 37.0
                & 30.9 & 61.9 & 42.3 & 36.0 & 29.5 & 15.9 \\
                DAnA $400$~\cite{chen2021should} & 52.0 & 83.3 & 59.6 & 42.9 & 60.0 & 45.4
                & 36.0 & 67.8 & 47.2 & 36.9 & 38.8 & 21.1 \\
                DAnA $800$~\cite{chen2021should} & 58.9 & 88.9 & 68.1 & 48.7 & 63.8 & 54.8
                & 44.2 & 78.0 & 59.0 & 45.2 & 42.5 & 30.3 \\
                \hline
                ODIP Stage $2$ (256 images)& 52.5 & 86.7 & 64.3 & 41.0 & 61.7 & 42.8 
                & 38.5 & 75.2 & 52.2 & 36.9 & 42.2 & 22.8\\
                ODIP Stage $4$ (512 images)& 54.6 & 88.8 & 66.3 & 46.3 & 62.2 & 43.5 
                & 40.8 & 79.7 & 57.9 & 42.3 & 47.1 & 21.2\\
                ODIP Stage $8$ (1024 images) & 59.8 & 91.4 & 65.6 & 51.5 & 65.9 & 56.2
                & 45.6 & 85.1 & 56.6 & 48.0 & 44.9 & 32.9 \\
                ODIP Stage $12$ (1536 images) & 59.9 & 91.8 & 66.1 & 49.8 & 66.0 & 56.2
                & 47.7 & 86.1 & 59.9 & 50.2 & 45.5 & 35.0 \\
                ODIP Stage $16$ (2048 images) & 62.4 & 91.7 & 69.6 & 53.5 & 67.5 & 59.2
                & 49.4 & 86.7 & 62.4 & 52.0 & 46.6 & 36.7 \\
                \bottomrule
        \end{tabular}
    }
    \caption{Quantitative analysis on the proposed YCB2D and Dense-YCB2D datasets. Our approach outperforms both the standard object detector (row 1-4) and state-of-the-art few-shot object detector (row 5-8). Note that both baseline methods require fully supervised data to learn, while our approach can learn from the interaction with a grasping system. At stage $4$, only $512$ images are collected but ODIP has achieved remarkable performance in comparison to Faster R-CNN using $2048$ fully-annotated images.
}
    \label{tab:stage}
\end{table*}
%

%
\section{EXPERIMENTS}

\subsection{Experiment Details}
In this work, the experiments are conducted in a robotic simulator to ensure reproducibility.
To test the generalization ability, we leverage the robotic simulator and YCB dataset~\cite{calli2015ycb} to construct a simulated 2D object detection dataset called YCB2D (see Fig.~\ref{fig:prediction}).
The proposed YCB2D is comprised of $4$ object categories ($e.g.$, cube, can, box, bottle).
In addition, we prepare a much more challenging dataset called Dense-YCB2D.
There will be at most $22$ dense objects of various classes in a Dense-YCB2D image.
%
%
%

%
We choose C2F~\cite{jeng2020coarse} and DAnA~\cite{chen2021should} as the grasp detector $g_{\theta}$ and few-shot object detector $f_{\theta}$ respectively due to their remarkable performance.
In the experiments, the maximum stage is $T=16$ and GOR interaction will be performed $N=16$ times for each individual category in each stage.
For instance, after stage $2$ there will be totally $128$ UDO images and $128$ MOA images stored in the ODIP database.
At training and inference, we provide FSOD models with $k=3$ support images to predict.
The $L$ and $\eta$ in Algorithm.~\ref{alg:algorithm1} are $16$ and $0.0001$ respectively.
All the baseline models and ODIP have been pre-trained on COCO 2017 benchmark. 
%
%

%
Instead of treating fine-tuning as an off-line process, ODIP can be adapted in an online manner where the $4$ object categories in YCB2D are sequentially introduced to the system.
The baseline methods~\cite{ren2015faster, chen2021should}, however, can only be fine-tuned in a traditional manner.
Therefore, we have to prepare a fine-tuning set for baseline methods in advance.
To generate the fine-tuning set, we compute the ground-truth bounding box annotations of all the objects by rendering the object coordinates based on the configurations of the simulator. 
The images in the fine-tuning dataset prepared for baselines are comprised of $4$ to $7$ multi-class objects and are fully annotated.
Though ODIP is trained in a more challenging scenario with incomplete annotations, it shows remarkable performance in comparison to baselines (Tab.~\ref{tab:stage}). 
%
%

%

%
%
\subsection{Quantitative Analysis}
In Tab.~\ref{tab:stage}, we compare ODIP with Faster R-CNN~\cite{ren2015faster} and the state-of-the-art FSOD model, DAnA~\cite{chen2021should}.
Faster R-CNN is fine-tuned on $256, 512, 1024$ and $2048$ images and DAnA is fine-tuned on $40, 200, 400$ and $800$ images with the conventional fine-tuning procedure.
Note that the data used to fine-tune Faster R-CNN and DAnA are fully annotated.
%
%

%
It can be observed in Tab.~\ref{tab:stage} that the performance of Faster R-CNN could be improved as more annotated data are given.
However, the increase in data only brings limited improvement, especially when testing on Dense-YCB2D.
Such results imply general object detectors could only be deployed to the environments consistent with the domain of training data.
If the model is to be applied to more cluttered environments, it might fail to identify target objects.
The few-shot object detectors are designed for efficient adaptation, and we observed DAnA indeed outperforms the general object detector as limited fine-tuning data are given.
However, it is obvious in Tab.~\ref{tab:stage} that its performance still heavily depends on the size of the fine-tuning set.
The model would still suffer overfitting if only a few annotated images are available.
%
%

%
Tab.~\ref{tab:stage} shows ODIP significantly outperforms the baselines, especially when testing on Dense-YCB2D.
At stage $4$, only $256$ UDO images and $256$ MOA images are collected by interacting with the grasping system.
However, in comparison to Faster R-CNN using $2048$ fully-annotated images, ODIP has surpassed Faster R-CNN on both YCB2D and Dense-YCB2D without ground-truth bounding boxes.
%
%

%

%
%
%

\begin{table}[t!]
    \centering
    \resizebox{\linewidth}{!}{
        \begin{tabular}{cc|cccc}
            \toprule
                \multirow{2}{*}{UDO data} & \multirow{2}{*}{MOA data} & \multicolumn{4}{c}{$AP$}\\ 
                \cline{3-6}
                && stage $2$ & stage $4$ & stage $6$ & stage $8$\\
                \hline
                \checkmark & & 21.2 & 33.0 & 32.8 & 32.7 \\
                & \checkmark & 18.4 & 25.0 & 31.9 & 38.6 \\
                \checkmark & \checkmark & 38.5 & 40.8 & 44.3 & 45.6\\
            \bottomrule
      \end{tabular}
    }
    \caption{The ablation study. Ideally, models should be trained on images comprised of dense, multi-class objects with ground-truth bounding box annotations, but such data are unavailable without human intervention. We propose a joint-training strategy to separate the required data into two different domains ($i.e.$, the dense-object UDO and the multi-class MOA images) that compensate each other.
    }
    \label{tab:ablation}
\end{table}
\subsection{Ablation Study}
As we discussed in Sec.~\ref{tab:disentangling}, the ideal training data should be multi-class, dense-object and fully annotated, yet such data are unavailable without human intervention.
Therefore, an expedient method would be training the model on both UDO (dense-object) and MOA (multi-class) images compensating each other to acquire comprehensive domain knowledge.
However, such a strategy remains a presumption that the domain knowledge could be learned from the two separate datasets.
Therefore, we conduct the ablations to verify the effectiveness of such a strategy.
Tab.~\ref{tab:ablation} shows training on either only UDO images or only MOA images leads to lower performance.
If we train the model on MOA only, the sparse objects would inhibit the model from tackling more complicated tasks. 
On the other hand, the noise in pseudo labels of UDO data would have negative impact on performance, so the system requires MOA data with one-shot ground truth to guide the network.
Based on the results, we conclude that jointly training on both UDO and MOA images can learn more comprehensive information and result in better performance.
%
%
%
%

%

%

\section{CONCLUSION}

We explore an object detector that can be adapted to novel domains by interacting with a grasping system in an online and automated manner.
The ability of the model can be enhanced as more visual observations are collected during the interaction.
In the experiments, we leverage a robotic simulator to simulate the scenarios that previously unseen objects are sequentially introduced to the system.
The results demonstrate that the system can generalize to novel domains without any human-annotated data, and the proposed model outperforms both traditional and few-shot object detectors.

\section*{ACKNOWLEDGMENT}
This work was supported in part by the Ministry of Science and Technology, Taiwan, under Grant MOST 110-2634-F-002-026 and Qualcomm Technologies, Inc. We benefit from NVIDIA DGX-1 AI Supercomputer and are grateful to the National Center for High-performance Computing.

\bibliographystyle{IEEEtran}
\bibliography{iros2021}

\begin{thebibliography}{10}
\providecommand{\url}[1]{#1}
\csname url@rmstyle\endcsname
\providecommand{\newblock}{\relax}
\providecommand{\bibinfo}[2]{#2}
\providecommand\BIBentrySTDinterwordspacing{\spaceskip=0pt\relax}
\providecommand\BIBentryALTinterwordstretchfactor{4}
\providecommand\BIBentryALTinterwordspacing{\spaceskip=\fontdimen2\font plus
\BIBentryALTinterwordstretchfactor\fontdimen3\font minus
  \fontdimen4\font\relax}
\providecommand\BIBforeignlanguage[2]{{%
\expandafter\ifx\csname l@#1\endcsname\relax
\typeout{** WARNING: IEEEtran.bst: No hyphenation pattern has been}%
\typeout{** loaded for the language `#1'. Using the pattern for}%
\typeout{** the default language instead.}%
\else
\language=\csname l@#1\endcsname
\fi
#2}}

\bibitem{demo2021video}
``Odip: Towards automatic adaption for object detection by interactive
  perception,'' \url{https://www.youtube.com/watch?v=1E4JGFjqZP0}.

\bibitem{ren2015faster}
S.~Ren, K.~He, R.~Girshick, and J.~Sun, ``Faster r-cnn: Towards real-time
  object detection with region proposal networks,'' \emph{arXiv preprint
  arXiv:1506.01497}, 2015.

\bibitem{lin2017focal}
T.-Y. Lin, P.~Goyal, R.~Girshick, K.~He, and P.~Doll{\'a}r, ``Focal loss for
  dense object detection,'' in \emph{Proceedings of the IEEE international
  conference on computer vision}, 2017, pp. 2980--2988.

\bibitem{fan2020few}
Q.~Fan, W.~Zhuo, C.-K. Tang, and Y.-W. Tai, ``Few-shot object detection with
  attention-rpn and multi-relation detector,'' in \emph{Proceedings of the
  IEEE/CVF Conference on Computer Vision and Pattern Recognition}, 2020, pp.
  4013--4022.

\bibitem{fan2020fgn}
Z.~Fan, J.-G. Yu, Z.~Liang, J.~Ou, C.~Gao, G.-S. Xia, and Y.~Li, ``Fgn: Fully
  guided network for few-shot instance segmentation,'' in \emph{Proceedings of
  the IEEE/CVF conference on computer vision and pattern recognition}, 2020,
  pp. 9172--9181.

\bibitem{chen2021should}
T.-I. Chen, Y.-C. Liu, H.-T. Su, Y.-C. Chang, Y.-H. Lin, J.-F. Yeh, and W.~H.
  Hsu, ``Should i look at the head or the tail? dual-awareness attention for
  few-shot object detection,'' \emph{arXiv preprint arXiv:2102.12152}, 2021.

\bibitem{jeng2020coarse}
K.-Y. Jeng, Y.-C. Liu, Z.~Y. Liu, J.-W. Wang, Y.-L. Chang, H.-T. Su, and
  W.~Hsu, ``A coarse-to-fine (c2f) representation for end-to-end 6-dof grasp
  detection,'' \emph{arXiv preprint arXiv:2010.10695}, 2020.

\bibitem{qin2020s4g}
Y.~Qin, R.~Chen, H.~Zhu, M.~Song, J.~Xu, and H.~Su, ``S4g: Amodal single-view
  single-shot se (3) grasp detection in cluttered scenes,'' in \emph{Conference
  on robot learning}.\hskip 1em plus 0.5em minus 0.4em\relax PMLR, 2020, pp.
  53--65.

\bibitem{xie2020self}
Q.~Xie, M.-T. Luong, E.~Hovy, and Q.~V. Le, ``Self-training with noisy student
  improves imagenet classification,'' in \emph{Proceedings of the IEEE/CVF
  Conference on Computer Vision and Pattern Recognition}, 2020, pp.
  10\,687--10\,698.

\bibitem{zoph2020rethinking}
B.~Zoph, G.~Ghiasi, T.-Y. Lin, Y.~Cui, H.~Liu, E.~D. Cubuk, and Q.~V. Le,
  ``Rethinking pre-training and self-training,'' \emph{arXiv preprint
  arXiv:2006.06882}, 2020.

\bibitem{calli2015ycb}
B.~Calli, A.~Singh, A.~Walsman, S.~Srinivasa, P.~Abbeel, and A.~M. Dollar,
  ``The ycb object and model set: Towards common benchmarks for manipulation
  research,'' in \emph{2015 international conference on advanced robotics
  (ICAR)}.\hskip 1em plus 0.5em minus 0.4em\relax IEEE, 2015, pp. 510--517.

\bibitem{iscen2019label}
A.~Iscen, G.~Tolias, Y.~Avrithis, and O.~Chum, ``Label propagation for deep
  semi-supervised learning,'' in \emph{Proceedings of the IEEE/CVF Conference
  on Computer Vision and Pattern Recognition}, 2019, pp. 5070--5079.

\bibitem{hinton2015distilling}
G.~Hinton, O.~Vinyals, and J.~Dean, ``Distilling the knowledge in a neural
  network,'' \emph{arXiv preprint arXiv:1503.02531}, 2015.

\bibitem{yalniz2019billion}
I.~Z. Yalniz, H.~J{\'e}gou, K.~Chen, M.~Paluri, and D.~Mahajan, ``Billion-scale
  semi-supervised learning for image classification,'' \emph{arXiv preprint
  arXiv:1905.00546}, 2019.

\bibitem{deng2009imagenet}
J.~Deng, W.~Dong, R.~Socher, L.-J. Li, K.~Li, and L.~Fei-Fei, ``Imagenet: A
  large-scale hierarchical image database,'' in \emph{2009 IEEE conference on
  computer vision and pattern recognition}.\hskip 1em plus 0.5em minus
  0.4em\relax Ieee, 2009, pp. 248--255.

\bibitem{karlinsky2019repmet}
L.~Karlinsky, J.~Shtok, S.~Harary, E.~Schwartz, A.~Aides, R.~Feris, R.~Giryes,
  and A.~M. Bronstein, ``Repmet: Representative-based metric learning for
  classification and few-shot object detection,'' in \emph{Proceedings of the
  IEEE/CVF Conference on Computer Vision and Pattern Recognition}, 2019, pp.
  5197--5206.

\bibitem{sung2018learning}
F.~Sung, Y.~Yang, L.~Zhang, T.~Xiang, P.~H. Torr, and T.~M. Hospedales,
  ``Learning to compare: Relation network for few-shot learning,'' in
  \emph{Proceedings of the IEEE Conference on Computer Vision and Pattern
  Recognition}, 2018, pp. 1199--1208.

\bibitem{ten2017grasp}
A.~ten Pas, M.~Gualtieri, K.~Saenko, and R.~Platt, ``Grasp pose detection in
  point clouds,'' \emph{The International Journal of Robotics Research},
  vol.~36, no. 13-14, pp. 1455--1473, 2017.

\bibitem{liang2019pointnetgpd}
H.~Liang, X.~Ma, S.~Li, M.~G{\"o}rner, S.~Tang, B.~Fang, F.~Sun, and J.~Zhang,
  ``Pointnetgpd: Detecting grasp configurations from point sets,'' in
  \emph{2019 International Conference on Robotics and Automation (ICRA)}.\hskip
  1em plus 0.5em minus 0.4em\relax IEEE, 2019, pp. 3629--3635.

\bibitem{ni2020pointnet++}
P.~Ni, W.~Zhang, X.~Zhu, and Q.~Cao, ``Pointnet++ grasping: learning an
  end-to-end spatial grasp generation algorithm from sparse point clouds,'' in
  \emph{2020 IEEE International Conference on Robotics and Automation
  (ICRA)}.\hskip 1em plus 0.5em minus 0.4em\relax IEEE, 2020, pp. 3619--3625.

\bibitem{murali20206}
A.~Murali, A.~Mousavian, C.~Eppner, C.~Paxton, and D.~Fox, ``6-dof grasping for
  target-driven object manipulation in clutter,'' in \emph{2020 IEEE
  International Conference on Robotics and Automation (ICRA)}.\hskip 1em plus
  0.5em minus 0.4em\relax IEEE, 2020, pp. 6232--6238.

\end{thebibliography}

\end{document}